# Challenges in bridging Social Semantics and Formal Semanticson the Web

Fabien Gandon, Michel Buffa, Elena Cabrio, Olivier Corby, Catherine Faron-Zucker, Alain Giboin, Nhan Le Thanh, Isabelle Mirbel, Peter Sander, Andrea Tettamanzi, Serena Villata

Inria, Univ. Nice Sophia Antipolis, CNRS, I3S, UMR 7271, 06900 Sophia Antipolis, France
`fabien.gandon@inria.fr`

**Abstract.** This paper describes several results of Wimmics, a research lab which names stands for: web-instrumented man-machine interactions, communities, and semantics. The approaches introduced here rely on graph-oriented knowledge representation, reasoning and operationalization to model and support actors, actions and interactions in web-based epistemic communities. The research results are applied to support and foster interactions in online communities and manage their resources.

## 1   Introduction: toward hybrid societies on the web

The Web is no longer perceived as a documentary system. Among its many evolutions, it became a virtual place where persons and software interact in mixed communities *i.e.* a hybrid space where humans and web robots interact and form new kinds of collectives that we will call here *hybrid societies*. These large scale interactions create many problems in particular the one of reconciling formal semantics of computer science (e.g. logics, ontologies, typing systems, etc.) on which the Web architecture is built, with soft semantics of people (e.g. posts, tags, status, etc.) on which the Web content is built.

Let us take a concrete and very common example. Many Web sites include forums, blogs, status feeds, wikis, etc. In other words, many Web sites include content management systems and rapidly build huge collections of information resources. As these collections grow, several tasks become harder to automate: search, notification, restructuring, navigation assistance, recommendation, trend analysis, etc. One of the main problems is the gap between the fairly informal way content is generated (e.g. plain text, short messages, free keywords) and the need for structured data and formal semantics to automate these functionalities (e.g. efficient indexes, domain thesauri). Mixed structures are starting to appear (e.g. structured folksonomies, hash tags, machine tags, etc.) but automating support in such collaboration spaces requires efficient and complete methods to fully bridge that gap.

As the Web becomes a ubiquitous infrastructure bathing all the objects of our world, this is just one example of the many frictions it will create between formal semantics and social semantics. This trend is also amplified by the growing number of datasets published and interlinked online by initiatives like Linking Open Data. This expanding web of data together with the schemas, ontologies and vocabularies used to structure and link it form a formal semantic web with which we have to design new interaction means to support the next generation of web applications.

This is why the Wimmics[1] research laboratory proposes to study methods, models and algorithms to bridge formal semantics and social semantics on the Web. This article provides a survey of the current research topics of the laboratory.

## 2    Challenges in bridging formal semantics and social semantics

From a formal modeling point of view one of the consequences of the evolutions of the Web is that the initial graph of linked pages has been joined by a growing number of other graphs. This initial graph is now mixed with sociograms capturing the social networks structure, workflows specifying the decision paths to be followed, browsing logs capturing the trails of our navigation, automatas of service compositions specifying distributed processing, linked open data from distant datasets, etc.

Moreover, these graphs are not available in a single central repository but distributed over many different sources with very different characteristics. Some sub-graphs are public (e.g. dbpedia) while others are private (e.g. corporate data). Some sub-graphs are small and local (e.g. a users' profile on a device), some are huge and hosted on clusters (e.g. Wikipedia).Some are largely stable (e.g. thesaurus of Latin), some change several times per second (e.g. sensor data), etc.

And each type of network of the Web is not an isolated island. Networks interact with each other: the networks of communities influence the message flows, their subjects and types, the semantic links between terms interact with the links between sites and vice-versa, the small changing graphs of sensors are joint to the large stable geographical graphs that position them, etc.

Not only do we need means to represent and analyze each kind of graphs, we also need the means to combine them and to perform multi-criteria analysis on their combination.

Wimmics proposes to address this problem focusing on the characterization of (a) typed graphs formalisms to model and capture these different pieces of knowledge and (b) hybrid operators to process them jointly. Our team especially considers the problems that occur in such structures when we blend formal stable semantic models and socially emergent and evolving semantics.

---

[1] Wimmics is a joint research laboratory between Inria Sophia Antipolis - Méditerranée and I3S (CNRS and Université Nice Sophia Antipolis). http://wimmics.inria.fr

The two next sections detail this research program according to two main research directions combining two complementary types of contributions we target:

1. First research direction: to propose multidisciplinary approach to analyze and model the many aspects of these intertwined information systems, their communities of users and their interactions;
2. Second research direction: to propose formalizations of the previous models and reasoning algorithms on these models providing new analysis tools and indicators, and supporting new functionalities and better management.

In a nutshell, the first research direction looks at models of systems, users, communities and interactions while the second research direction considers formalisms and algorithms to represent them and reason on their representations.

In the short term we intend to survey, extend, formalize and provide reasoning means over models representing systems, resources, users and social links in the context of social semantic web applications.

In the longer term we intend to extend these models (e.g. dynamic aspects), unify their formalisms (dynamic typed graphs) and propose mixed operations (e.g. metrical and logical reasoning) and algorithms to scale them (e.g. random walks) to support the analysis of epistemic communities' structures, resources and dynamics.

Ultimately our goal is to provide better collective applications on the web of data and the semantic web addressing jointly two sides to the problem: (1) improve access and use of the linked data for epistemic communities and at the same time (2) use typed graph formalisms to represent the web resources, users and communities and reason on them to support their management.

## 3  Analyzing and modeling users, communities and their interactions in a Social Semantic Web context

The first challenge we identified in the introduction was to propose multidisciplinary approach to analyze and model intertwined information systems, communities of users and their interactions. Examples of research questions here include:

- How do we improve our interactions with an information system that keeps getting more and more complex?
- How do we reconcile and integrate the formalized stable semantics of computer science and the negotiable social interactions?
- How do we facilitate communication between systems and system developers using formal representations and between users and usage analysts using corresponding less formal or non-formal representations?
- How do we reconcile local contexts of users and global characteristics of the world-wide system?

The main goal of this first research direction is to improve the understanding the systems have of the communities of their users. To provide better collective applica-tions on the web of data and the semantic web we need to adapt classical models to the specificities and variety of web systems (requirements, functionalities, specifications), users (profiles and context: location, devices, activities, etc.), and groups (communities and networks of interest, communities and networks of practice, social constructs, etc.). In Wimmics, these models are designed, integrated, and published according to web standards to support functionalities of web applications providing access and use of the linked data to epistemic communities.

We defend that such models necessarily call for multidisciplinary approaches to analyze and model the many aspects of the information systems.Our proposal relies:

— on extending requirement modeling to build models of the systems;
— on cognitive studies and user modeling results to build models of the users;
— on ergonomic studies and interaction design methods to model the interactions between users through the system and with the system, in order to support and improve these interactions.

We had several experiences in past projects with a user modeling technique known as Personas [1]. We are interested in these user models that are represented as specific, individual humans and we apply them to capture models of the members of web-based communities. Personas are derived from significant behavior patterns (i.e., sets of behavioral variables) elicited from interviews with and observations of users (and sometimes customers) of the future product. The main merit of the Personas method is to engage design team members more effectively in not only taking users into account but also having constantly in mind that they are designing for people. This effectiveness comes from several aspects, in particular:

1. by integrating concrete elements in the description of a user-type (name, photo, etc.)., it prevents that the user remains an abstraction for the designer (this abstraction leading the designer to lose sight of the user);
2. by connecting more strongly scenarios to the actors of these scenarios (i.e., the users), the Personas method avoids the problem often encountered in conventional methods of scenario-based design, namely to ignore users in the development of scenarios and thus "dehumanizing" these scenarios.

In the Personas method, the link between scenarios and users is established with the most important characteristic of personas: their goals.These goals form the basis for scenario development. Our user models specialize "Personas" approaches to include aspects appropriate to Web applications. The formalization of these models will rely on ontology-based modeling of users and communities starting with generalist schemas (e.g. FOAF).

Beyond the individual user models we propose to rely on social studies to build models of the communities, their vocabularies, activities and protocols in order to identify where and when formal semantics is useful. We already proposed an extension of the Persona approach to Collective Personas [2] and wenow develop our method to encompass web-based communities. We compare this approach to the related "collaboration personas" method [3][4] and to the group modeling methods [5] which are extensions to groups of the classical user modeling techniques dedicated to individuals. Both methods having been developed independently from one another, they differ along several dimensions, for example: whereas the Collaboration Personas method (CnP) focuses on forms of collaboration within a group, the Collective Personas method (CeP) focuses on the nature of the collective; whereas CeP refers to models or theories of collectives to define types of collectives, CnP uses pragmatic classifications; whereas CnP describes scenarios as stories, CeP describes them in a more structured way. We also propose to rely on and adapt participatory sketching and prototyping to support the design of interfaces for visualizing and manipulating representations of collectives.

Wimmics also has a background in requirement models and, in the short term, we want to consider their extension and specialization to web applications (in particular semantic web and linked data applications) and their representation in web-based formalisms in order to support mutual understanding and interoperability between requirements, resources and specifications of interconnected web applications and web datasets. (e.g. [6])

For all the models we identify we rely on and evaluate knowledge representation methodologies and theories, in particular ontology-based modeling.

In addition to the persona models identified previously, in a longer term we will consider a number of additional features to be captured in the user models.

We already added contexts, devices, processes and media descriptions that are formalized and used to support adaptation, proof and explanation and foster acceptation and trust from the users [46][47]. We specifically target a unified formalization of these contextual aspects to be able to integrate them at any stage of the processing. This unified formalization already allows us to use the same model and data for very different functionalities such as access control [7] or presentation customization [8].

We extended current descriptions of relational and emotional aspectsin existing variants of the personas technique. In particular we intendto exploit Olsen's characterization of a user's relationship to a product [9] since this characterization was made by analogy to human relationships. The elaboration of the characteristics to be included in a "relational-user" model will rely on work dealing with "relational agency" ("as a capacity to recognize others as resources, to elicit their interpretations and to negotiate aligned actions" [10]), and on "relational agents" (as "computational artifacts designed to build long-term social-emotional relationships with users" [11]). The elaboration of relational characteristics will be informed by empirical studies of relationship characterization, and of personality definition by users on the Web. These directions are a natural extension of our work on the use of affective ontologies in semantic web applications [12] and algebraic modeling of emotional states[13]. We also proposed algorithm to detect and classify the emotional states [14][15] that can then be represented and exchanged together with their ontologies on the Web. Indeed, for each of these extensions we systematically consider additional extensions of the corresponding schemas to capture additional aspects and publish them as public ontologies on the semantic web.

Concerning the social dimension we now focus on studying and modeling mixed representations containing social semantic representations (e.g. folksonomies) and formal semantic representations (e.g. ontologies) and propose operations that allow us to couple them and exchange knowledge between them[16]. The very long term objective is to obtain a uniformed and integrated representation of social aspects (e.g., groups, networks, communities), social objects (web resources e.g. pictures, posts), social informal semantics (e.g. tags, folksonomies) and social formal semantics (e.g. ontologies, schemas).

In addition, to take into account social dynamics, we believe that argumentation theory can provide models (e.g. [17]) that must be adapted to open web constraints. Argumentation theory can be combined to requirement engineering to improve participant awareness and support decision-making (e.g. [18]). On the methodological side, we propose to adapt to the design of such systems the incremental formalization approach originally introduced by the CSCW and HCI communities [19][20]. In incremental formalization users first express information informally and then the system helps them formalize it. The goal of such an approach is to get users to interact at least partially with formal representations, to make them contribute to a formalization closer to their needs [21].

Argumentation theory can also be combined with semantic web models and social web approaches to provide explicit formal representation of some social dynamics (e.g. opinions, agreements, debates, disagreements) more and more useful to understand the state and status of a resource and, for instance, decide on whether to trust it or not.

This kind of understanding and models allow scaling-up some very time-consuming tasks on the social web (e.g. managingand moderating Wikipedia) in particular when they are combined with natural language processing (NLP) to tackle the textual nature of these interactions as in [22][23][24][25][26].In addition NLP approaches can also improve the design of interaction with the collections of models and data that are growing on the Web. For instance NLP allows us to support natural language querying of semantic web triple stores [27][28].

And the linguistic knowledge useful for such processing can in turn be the subject of knowledge representation and data exchanged on the Web [29][30].

Finally, on a very long term a much needed evolution of all our models is the temporal and dynamic dimension. We now plan to study and survey initiatives and contributions on dynamic graph representation and analysis and merge them with typed graph models of the Web of data to natively and uniformly support the time dimension in our representations.

## 4   Formalizing and reasoning on heterogeneous semantic graphs

The second challenge identified in the introduction is to propose formalizations of the previous models and reasoning algorithms on these models providing new analysis tools and indicators, and supporting new functionalities and better management.Examples of research questions include for instance:

- What kind of formalism is the best suited for the models of the previous section?
- How do we analyze graphs of different types and their interactions?

— How do we support different graph life-cycles, calculations and characteristics in a coherent and understandable way?

In this second research direction members of Wimmics focus on formalizing as typed graphs the models identified in the previous section in order for software to exploit them in their processing. The challenge is two-sided:

— To propose models and formalisms to capture and merge representations of both kinds of semantics (e.g. formal ontologies and social folksonomies). The important point is to allow us to capture those structures precisely and flexibly and yet create as many links as possible between these different objects.
— To propose algorithms (in particular graph-based reasoning) and approaches (e.g. human-computing methods) to process these mixed representations. In particular we are interested in allowing cross-enrichment between them and in exploiting the life cycle and specificities of each one to foster the life-cycles of the others.

While some of these problems are known, for instance in the field of knowledge representation and acquisition (e.g. disambiguation, fuzzy representations, and argumentation theory), the Web reopens them with exacerbated difficulties of scale, speed, heterogeneity, and an open-world assumption by default.

Many approaches emphasize the logical aspect of the problem especially because logics are close to computer languages. We defend that the graph nature of linked data on the Web and the large variety of types of links that compose them call for typed graphs models. We believe the relational dimension is of paramount importance in these representations and we propose to consider all these representations as fragments of a typed graph formalism directly built above the semantic Web formalisms. Our choice of a graph based programming approach for the semantic and social web and of a focus on one graph based formalism is also an efficient way to support interoperability, genericity, uniformity and reuse.

We targeted an abstract graph model close to the GRIWES model [31] and we evaluate it in merging social graphs (e.g. sociograms, folksonomies) and semantic Web graphs (e.g. RDF, schemas, linked data) in a unified typed graph formalism. This work on abstracting the knowledge representation models follows our experience with conceptual graphs and semantic networks approaches.

An example of such abstract structure is the ERGraph[31] defined relatively to a set of labels $L$ as a 4-tuple $G=(E_G, R_G, n_G, l_G)$ where:

— $E_G$ and $R_G$ are two disjoint finite sets respectively, of nodes called entities and of hyperarcs called relations.
— $n_G : R_G \rightarrow E_G^*$ associates to each relation a finite tuple of entities called the arguments of the relation.
— $l_G: E_G \cup R_G \rightarrow L$ is a labelling function of entities and relations.

This type of oriented labelled multi-graph structure is at the core of most of our formalizations. New knowledge structures are regularly identified (e.g. folksonomies, named graphs) and old ones re-launched (e.g. thesauri and SKOS). This kind of ab-stract construct can be used and reused across graph representations such as RDF, Topic Maps, Social Networks, Knowledge Graphs, etc.

There exists now an extensive body of work in Graph-based Knowledge Representation [32] that we align with the ones needed for semantic web data structures (e.g. [31][33][34]) and in the short term we intend to continue specifying the required characteristics of such a language and systematically evaluate their effectiveness in implementing these abstract graph models in real applications [35][36].

Likewise we extend our abstract graph machine not only to cover as many features as possible of new languages like SPARQL 1.1, RDF 1.1 and RIF, but also to extend them with experimental features (e.g. semantic distances) and a challenge is to integrate other operators with classical graph manipulation in particular: approximation, clustering, analysis operations, spreading algorithm, temporal reasoning and extend them to work on typed graphs. For instance, we considered the near linear time algorithm to detect community structures in large scale network RAK/LP [37] based on label propagation and changed the propagation algorithm to take semantics into account the algorithmSemTagP[38].

Currently, graph operators (joint, homomorphism, propagation, distances, etc.) allow us to perform a broad range of queries and reasoning operations. An example of abstract graph operation is an ERMapping[31]: Let $G$ and $H$ be two ERGraphs, an ERMapping$_{<x>}$ from $H$ to $G$ for a binary relation $X$ over $L \times L$ is a partial function $M$ from $E_H$ to $E_G$ such that:

— $\forall e \in M^{-1}(E_G), (l_G(M(e)), l_H(e)) \in X$
— $\forall r' \in R_{H'} \exists r \in R_G$ such that $card(n_{H'}(r')) = card(n_G(r))$
— $\forall\ 1 \leq i \leq card(n_G(r)), M(n_{H'}^{i}(r')) = n_G^{i}(r)$
— $\forall r' \in R_{H'} \exists r \in M(r')$ such that $(l_G(r), l_H(r')) \in X$ where $H'$ is the sub-ERGraph of H induced by $M^{-1}(E_G)$.

This mapping operator can then be used and reused for many operations (searching, deriving, grouping, etc.) and across many graph formalisms compatible with the ERGraph structure. In particular when $X$ is a preorder over $L$, it captures a hierarchy such as the taxonomical skeleton of an ontology, a thesaurus, a partonomy, etc.

Our implementations [35] and their applications (e.g.ISICIL[39], DATALIFT[40], DiscoveryHub[41]) show that type-based inference algorithms (e.g. conceptual graph projection, inference rules) and type-parameterized operators (e.g. parameterized betweeness centrality) provide declarative formalisms to flexibly define operations to monitor, filter, query, mine, validate, protect, etc. these imbricated graph structures taking into account constraints spanning several types of network at once.

In the longer term we intend to build on our experience with such formalisms to identify, propose and characterize fragments of typed graph formalisms best suited for each type of model identified before. We will restrain ourselves to specify the required characteristics of a limited number of formalisms (ideally one) and systematically evaluate their effectiveness in implementing these abstract graph models in real applications.

The mixed representations identified in the previous sectionreally call for hybrid reasoning methods merging semantic Web inferences, social graph analysis and content mining in cross-dimensional indicators and operators. The key problem is to have integrated operators on these formalisms, able to perform at the same time exact reasoning and more approximate one to combine all aspects of the problem. For instance a centrality [42] can be computed on a social network taking into account only some relation types [43] or some topics of interest using an extension of regular expressions to graph paths [33]. This same centrality can also be computed by using a complete walk algorithm or approximated by using random walks but in both cases the ability to consider and reason on types of links and nodes will be a core problem. Another example is combining folksonomies and ontologies in the same application and using a combination of automated processing (e.g. a range of semantic similarities and inference rules) and human-based computing (e.g. analyzing the behaviors of users carrying out a search) to structure and maintain a thesaurus of tags and keywords used in a community [44].

Our final goal is to have both an abstract language dissociated from the concrete languages and an extensible abstract machine to process them. In particular this allows us to define parameterizable graph operators for instance to revisit classical structural metrics and adapt their definition to go beyond the pure structural calculation and take into account the types in the graphs. In the longer term, we intend to perform search (e.g. homomorphism) and logical derivation (e.g. homomorphism and merge) but also approximation (e.g. distances), clustering (e.g. propagation), analysis (e.g. centrality), etc. jointly on the same graphs. We target the design of an abstract graph machine [35] generalizing operations needed by and sometime shared across different languages (e.g. SPARQL, RIF, POWDER, RDF/S and OWL inferences) and operations. In addition we also believe it is interesting to study alternatives to OWL stack and the associated DL-reasoning. For instance a rule-based semantic web with an alternative stack (RDF/S + SPARQL + Rules) provides certain advantages: rules are often more natural for humans, they support event-based programming and web service integration, they are usable both for domain independent and domain dependent inferences, etc.

To adapt to web growth and dynamics, we also plan to evaluate other approaches (e.g. Random walk on graphs approaches) that do not naturally use labels but could be indirectly parameterized (e.g. making a correspondence between probabilistic distributions and the types of links) and to consider temporal reasoning approaches to include temporal context and change patterns to identify trends, mine temporal propagation to build oriented networks, track behavioral patterns to qualify actors and communities (e.g. detect a dying community) extending models from [45] for instance.

We believe that moving to graph languages with open-world logics, temporal aspects, distributed and loosely coupled algorithms and model-driven programming relying on higher abstractions (e.g. formal ontologies) provides an adequate theoretical and operational framework to allow not only the specification and operationalization of the models and algorithms,but also the opening of these black boxes to be able to explain, document, prove and trace query performances and results for the users, as in [46] and [47].

As a last example, the same graph-based formalisms we propose to use in representing our models can be used to declaratively capture the landscape of the data distribution and the workflows of our operations [48], interpret them and execute them. This is the approach we would like to explore to support different operations on heterogeneous and distributed data (e.g. summarizing the content of distributed triple stores [49]) and automated explanation, trace and documentation of the processes.

## 5    Conclusion

Wimmics provides models and algorithms to bridge formal semantics and social semantics by formalizing and reasoning on heterogeneous semantic graphs.

The models we design include: users, their profiles, their requirements, their activities and their contexts; social links, social structures, social exchanges and processes; conceptual models including ontologies, thesauri, and folksonomies. Whenever possible these models are formalized and published according to standardized web formalisms and may motivate research and suggestions on extending these standards. The schemas and datasets produced are published as linked data following the web architecture principles.

The algorithms we study include: typed graphs indexing, reasoning and searching; hybrid processing merging logical inferences, rules and metrical inferences; approximation and propagation algorithms; distributed and scalable alternatives to classical reasoning. These algorithms are implemented and distributed as part ofgeneric open source software.

Wimmicshasindeed the culture of producing prototypes of applications and extensions of existing applications relying on web languages as demonstrators and proofs of concept. At the core of many of our prototypes is the abstract graph library we develop, maintain and publish as open-source software. This platform called Corese/KGRAM[35] currently implements W3C standards (in particular RDF/S 1.1 and SPARQL 1.1) and is both a research result and a library on top of which we test new ideas and algorithms. Currently this platform is extensively used in several applications such as: Isicil, Neurolog,DiscoveryHub, etc.

Finally we continuously participate to the extension, specifications, implementation, tests, deployment and teaching of W3C Web standards and our research results support, use and influence the evolution of these standards.

# 6 References

1. Cooper, A., Reinmann, R. (2003). About Face 2.0: The Essentials of Interaction Design. Indianapolis, IN: Wiley Publishing, Inc.
2. Alain Giboin. From Individual to Collective Persona: Modeling Realistic Groups and Communities of Users (and not Only Realistic Individual Users). Proc. of the Fourth International Conference Advances in Human-Computer Interactions (ACHI 2011), Gosier, Guadeloupe, February 23-28, 2011
3. Tara Matthews, Stephen Whittaker, Thomas Moran, Meng Yang. "Collaboration personas: A framework for understanding & designing collaborative workplace tools." In Workshop "Collective Intelligence In Organizations: Toward a Research Agenda." at Computer Supported Cooperative Work (CSCW), 2010.
4. Tara Matthews, Steve Whittaker, Thomas Moran, Sandra Yuen. (2011) "Collaboration personas: A new approach to designing workplace collaboration tools." In Proceedings of the SIGCHI Conference on Human Factors in Computing Systems (CHI), pp. 2247-2256.
5. E. Gaudioso, A. Soller, and J. Vassileva (Eds). User Modeling to Support Groups, Communities and Collaboration. User Modeling and User-Adapted Interaction 16, Special Issue, 2006.
6. Pascal Neveu and Caroline Domerg and Juliette Fabre and Vincent Nègre and Emilie Gennari and Anne Tireau and Olivier Corby and Catherine Faron-Zucker and Isabelle Mirbel, Using Ontologies of Software: Example of R Functions Management Proc. Third International Workshop on REsource Discovery, RED 2010, 2011, LNC
7. Costabello L., Villata S., Gandon F. Context-Aware Access Control for RDF Graph Stores. 20th European Conference on Artificial Intelligence, ECAI 2012
8. Luca Costabello, Error-Tolerant RDF Subgraph Matching for Adaptive Presentation of Linked Data on Mobile, ESWC, 2014
9. Olsen, G. (2004). Persona Creation and Usage Toolkit, Information Architecture (IA) Summit, 15 pages. http://www.asis.org/~iasummit/2004/finalpapers/Olsen_Handout_or__final__paper.pdf
10. Edwards, A. (2007) Relational Agency in Professional Practice: a CHAT analysis, Actio. An International Journal of Human Activity. 1. 1-17.
11. Bickmore, T.W., Picard, R.W. (2005). ACM Transactions on Computer-Human Interaction, Vol. 12, No. 2. (June 2005), pp. 293-327.
12. Alain Giboin, Motivating the Use of Affective Ontologies in Semantic Web Applications Which Mediate Interactions Between Members of Organizations or Communities, COOP'2008 Workshop on Affective aspects of cooperative interactions, Carry-Le-Rouet, May 2008.
13. ImenTayari, Nhan Le Thanh, Chori Ben Amar. "Towards an algebraic modeling of emotional states" in Proceedings of the The Fifth International Conference on Internet and Web Applications and Services (ICIW 2010), IARIA: Int. Academy, Research and Industry Association, pages 513-518, CPS-IEEE CSDL, Barcelona, Spain, 9-15 may 2010
14. Franck Berthelon and Peter Sander, Regression algorithm for emotion detection, CogInfoCom, IEEE International Conference on Cognitive Infocommunications, Budapest, 2013
15. Franck Berthelon and Peter Sander, Emotion Ontology for Context Awareness, CogInfoCom, IEEE International Conference on Cognitive Infocommunications, Budapest, 2013

16. Freddy Limpens, Alexandre Monnin, David Laniado, Fabien Gandon "NiceTag Ontology: tags as named graphs", International Workshop in Social Networks Interoperability, ASWC09, 2009
17. Phan Minh Dung: On the Acceptability of Arguments and its Fundamental Role in Non-monotonic Reasoning, Logic Programming and n-Person Games. Artificial Intelligence, 77(2), pp. 321-357, 1995
18. Isabelle Mirbel, Serena Villata, Enhancing Goal-based Requirements Consistency: an Argumentation-based Approach, 13th International Workshop on Computational Logic in Multi-Agent Systems (CLIMA 2012), LNCS Springer, Montpellier, France, August 2012.
19. Blythe, J., & Gil, Y. 2004. Incremental Formalization of Document Annotations through Ontology-based Paraphrasing. In Proc. of WWW'04 Thirteenth International World Wide Web Conference (May 17 - 22, 2004), 2004, New York, NY, USA.
20. Shipman III, F.M., & McCall, R. 1994. Supporting Knowledge-Base Evolution with Incremental Formalization. In Proceedings of CHI '94 (April 24-28, 1994), Boston, Mass., USA, pp. 285-291.
21. Giboin, A., Prié, Y. (2011). Interagir avec des représentations formelles. In Actes-complémentaires de la conférence IHM'2011, 57-58.
22. Elena Cabrio, Sara Tonelli and Serena Villata. A Natural Language Account for Argumentation Schemes, to appear in Proceedings of the XIII Conference of the Italian Association for Artificial Intelligence (AI*IA 2013). Turin, Italy, December 2013.
23. Elena Cabrio, Julien Cojan, Serena Villata and Fabien Gandon. Hunting for Inconsistencies in Multilingual DBpedia with QAKiS, to appear in Proceedings of the 12th International Semantic Web Conference (ISWC 2013). Poster/Demo paper. Sydney, Australia, October 2013.
24. Elena Cabrio, Julien Cojan, Serena Villata and Fabien Gandon. Argumentation-based Inconsistencies Detection for Question-Answering over DBpedia, to appear in Proceedings of the ISWC 2013 workshop NLP &DBpedia. Sydney, Australia, October 2013.
25. Elena Cabrio, Sara Tonelli, Serena Villata. From Discourse Analysis to Argumentation Schemes and back: Relation and Differences, to appear in Proceedings of the 14th International Workshop on Computational Logic in Multi-Agent Systems (CLIMA 2013). Corunna, Spain, September 2013.
26. Elena Cabrio, Serena Villata and Fabien Gandon, A Support Framework for Argumentative Discussions Management in the Web, Best Paper of the Extended Semantic Web Conference (ESWC 2013). Montpellier, France, May 2013.
27. Elena Cabrio, Julien Cojan, Fabien Gandon, and Amine Hallili. Querying Multilingual DBpedia with QAKiS, Proceedings of the Extended Semantic Web Conference (ESWC 2013). Demo paper. Montpellier, France, May 2013.
28. Julien Cojan, Elena Cabrio, Fabien Gandon, Filling the Gaps Among DBpedia Multilingual Chapters for Question Answering, Proceedings of ACM Web Science 2013. Paris, France, May 2013
29. MaximeLefrançois, RomainGugert, Fabien Gandon, and Alain Giboin, Application of the unit graphs framework to lexicographic definitions in the RELIEF project, to appear in Proc. of the 6th International Conference on Meaning-Text Theory (MTT 2013). Prague, Czech Republic, August 2013.
30. MaximeLefrançois, Fabien Gandon, Reasoning with dependency structures and lexicographic definitions, to appear in Proc. of the 2nd International Conference on Dependency Linguistics (Depling 2013). Prague, Czech Republic, August 2013.
31. Jean-François Baget, Olivier Corby, Rose Dieng-Kuntz, Catherine Faron-Zucker, Fabien Gandon, Alain Giboin, Alain Gutierrez, Michel Leclère, Marie-Laure Mugnier, Rallou-Thomopoulos. Griwes: Generic Model and Preliminary Specifications for a Graph-Based Knowledge Representation Toolkit. Proc. of the 16th International Conference on Conceptual Structures (ICCS'2008) July 2008 Toulouse

32. Michel Chein, Marie-Laure Mugnier, Graph-based Knowledge Representation: Computational Foundations of Conceptual Graphs (Advanced Information and Knowledge Processing), Springer, ISBN-13: 978-1849967693, 2009
33. Olivier Corby, Web, Graphs & Semantics, Proc. of the 16th International Conference on Conceptual Structures (ICCS'2008), July 2008 Toulouse
34. Fabien Gandon, Graphes RDF et leur Manipulation pour la Gestion de Connaissances, Habilitation à Diriger les Recherches (HDR), soutenue le Mercredi 5 novembre 2008.
35. Olivier Corby and Catherine Faron-Zucker, The KGRAM Abstract Machine for Knowledge Graph Querying, IEEE/WIC/ACM International Conference, September 2010, Toronto, Canada.
36. Corby O., Faron-Zucker C. (2007), Implementation of SPARQL Query Language based on Graph Homomorphism, In Proc. of the 15th International Conference on Conceptual Structures, ICCS 2007, LNCS 4604, Springer, Sheffield, UK
37. Raghavan, R.N., Albert, R., Kumara, S.: Near Linear Time Algorithm to Detect Community Structures in Large Scale Network. Phys. Rev. E, 76, 036106. (2007)
38. Guillaume Erétéo, Fabien Gandon, and Michel Buffa, SemTagP: Semantic Community Detection in Folksonomies, IEEE/WIC/ACM International Conference on Web Intelligence, August 2011, Lyon.
39. Michel Buffa, Guillaume Ereteo, Freddy Limpens, Fabien Gandon, Folksonomies and Social Network Analysis in a social Semantic Web, 39th International Conference on Current Trends in Theory and Practice of Computer Science, January 26–31, 2013
40. François Scharffe, GhislainAtemezing, RaphaëlTroncy, Fabien Gandon, Serena Villata, Bénédicte Bucher, FayçalHamdi, Laurent Bihanic, Gabriel Képéklian, Franck Cotton, JérômeEuzenat, Zhengjie Fan, Pierre-Yves Vandenbussche, Bernard Vatant, Enabling linked data publication with the Datalift platform, Proc. AAAI workshop on semantic cities, 2012
41. Nicolas Marie, MyriamRibiere, Fabien Gandon, FlorentinRodio, Discovery Hub: on-the-fly linked data exploratory search, to appear in Proc. of I-Semantics 2013
42. Freeman, L.C.: Centrality in social networks: Conceptual Clarification. Social Networks. 1, 215-239. (1979)
43. Guillaume Erétéo, Michel Buffa, Fabien Gandon, and Olivier Corby. Analysis of a Real Online Social Network using Semantic Web Frameworks. In Proc. International Semantic Web Conference, ISWC'09, Washington, USA, October 2009
44. Freddy Limpens, Fabien Gandon and Michel Buffa, Helping Online Communities to Semantically Enrich Folksonomies, Web Science Conference, April, 2010, Raleigh, NC, USA.
45. Angeletou, Sofia; Rowe, Matthew and Alani, Harith (2011). Modelling and analysis of user behaviour in online communities. In: 10th International Semantic Web Conference (ISWC 2011), 23 - 27 Oct 2010, Bonn, Germany
46. Rakebul Hasan, Generating and Summarizing Explanations for Linked Data, to appear in Proc. of 11th Extended Semantic Web Conference (ESWC2014)
47. Rakebul Hasan, Predicting SPARQL Query Performance and Explaining Linked Data, to appear in the PhD Symposium of 11th Extended Semantic Web Conference (ESWC2014)
48. Moussa Lo, Fabien Gandon, Semantic web services in corporate memories, ICIW 2007, International Conference on Internet and Web Applications and Services, May 13-19, 2007, Mauritius

49. Adrien Basse, Fabien Gandon, Isabelle Mirbel and Moussa Lo, Frequent Graph Pattern to Advertise the Content of RDF Triple Stores on the Web, Proc. Web Science Conference, Raleigh, NC, USA, April 2010.